\documentclass[lettersize,journal]{IEEEtran}
\usepackage{amsmath,amsfonts}
\usepackage{algorithmic}
\usepackage{algorithm}
\usepackage{array}
\usepackage[caption=false,font=normalsize,labelfont=sf,textfont=sf]{subfig}
\usepackage{textcomp}
\usepackage{stfloats}
\usepackage{url}
\usepackage{verbatim}
\usepackage{graphicx}
\usepackage{cite}

\usepackage{hyperref}       
\usepackage{booktabs}
\usepackage{tikz}
\usepackage{placeins} 
\begin{document}

\setlength{\abovecaptionskip}{5pt}  
\setlength{\belowcaptionskip}{5pt} 
\setlength{\extrarowheight}{0pt}   
\setlength{\textfloatsep}{5pt}     

\title{SatMamba: Development of Foundation Models for Remote Sensing Imagery Using State Space Models}

\author{\IEEEauthorblockN{Chuc Man Duc, Hiromichi Fukui}
\thanks{Chuc Man Duc is with Department of Computer Science, Faculty of Information Technology, University of Engineering and Technology, Vietnam National University, Hanoi, Vietnam; Hiromichi Fukui are with International Digital Earth Applied Science Research Center, Chubu University, Kasugai, Aichi, 487-8501, Japan (e-mail: chucmd@vnu.edu.vn, fukui@isc.chubu.ac.jp).}}


\markboth{Journal of \LaTeX\ Class Files,~Vol.~14, No.~8, September~2024}%
{Shell \MakeLowercase{\textit{et al.}}: A Sample Article Using IEEEtran.cls for IEEE Journals}


\maketitle

\begin{abstract}
  Foundation models refer to deep learning models pretrained on large unlabeled datasets through self-supervised algorithms. In the Earth science and remote sensing communities, there is growing interest in transforming the use of Earth observation data, including satellite and aerial imagery, through foundation models. Various foundation models have been developed for remote sensing, such as those for multispectral, high-resolution, and hyperspectral images, and have demonstrated superior performance on various downstream tasks compared to traditional supervised models. These models are evolving rapidly, with capabilities to handle multispectral, multitemporal, and multisensor data. Most studies use masked autoencoders in combination with Vision Transformers (ViTs) as the backbone for pretraining. While the models showed promising performance, ViTs face challenges, such as quadratic computational scaling with input length, which may limit performance on multiband and multitemporal data with long sequences. This research aims to address these challenges by proposing SatMamba, a new pretraining framework that combines masked autoencoders with State Space Model, offering linear computational scaling. Experiments on high-resolution imagery across various downstream tasks show promising results, paving the way for more efficient foundation models and unlocking the full potential of Earth observation data. The source code is available in https://github.com/mdchuc/HRSFM.
\end{abstract} 


\section{Introduction}
\IEEEPARstart{F}{oundation} models have received significant attention in the field of remote sensing (RS) in recent years. These deep learning models are pretrained on large datasets and can be fine-tuned for various downstream tasks. They have been reported to achieve state-of-the-art results in key RS applications using medium-resolution satellite imagery, such as land-cover classification/semantic segmentation, scene classification, and change detection \cite{Cong2022,Jakubik,Hong2023,Xiong}. Foundation models are often trained using masked image modeling, a self-supervised learning (SSL) approach. In this framework, part of the input image is masked, and the model is trained to predict the masked portion \cite{He2022}. This approach is widely used to train foundation models in RS due to its straightforward procedure for preparing pretraining data. Specifically, foundation models are built as Masked Autoencoders (MAE), using the Vision Transformer (ViT) \cite{Dosovitskiy2021} as the backbone. The MAE has been demonstrated to be a scalable learner, capable of learning in a self-supervised manner from vast amounts of unlabeled data. 

Currently, the ViT is the predominant backbone in RS foundation models. However, with the recent emergence of state-space models \cite{Gu2023}, which have been successfully applied in fields like medical imaging, semantic segmentation in natural images \cite{Ruan,Zhu2024,Liu2024,Chen2024}, and satellite imagery, the question of whether and how to build a foundation model using Mamba as the backbone remains an open research challenge. Moreover, the fine-tuning process plays a critically important role, particularly in image-to-image tasks. Notably, variations in image size and normalization methods between pretraining and fine-tuning phases can significantly influence fine-tuning performance \cite{Corley2023}. Recent studies have reported that RS foundation models based on MAE and ViT, such as SatMAE and Prithvi, may perform less effectively than a basic UNet model on image-to-image downstream tasks \cite{Fibaek2024}.

In this research, we propose SatMamba, a novel pretraining architecture that integrates Mamba into a masked autoencoder framework, demonstrating competitive performance compared to ViT-based architectures across various image-to-image downstream tasks. While SatMamba is primarily evaluated on high-resolution data, it has the potential to work with other image domains, including multispectral, medium-resolution remote sensing, and natural images, positioning it as a viable alternative to ViT-based pretraining architectures.

\section{Background}
\label{sec:headings}
\subsection{Masked Autoencoder}
The Masked Autoencoder \cite{He2022} is a widely used architecture for pretraining foundation models in the remote sensing domain. In its basic form, the MAE processes images $I \in \mathbb{R}^{C \times H \times W}$, where \textit{C}, \textit{H}, \textit{W} represent the number of spectral bands, height, and width of the image, respectively. Through patch embedding, the input image \textit{I} is transformed, typically by a 2-D convolutional layer with a non-overlapping sliding window, into a sequence of 1-D vectors, also known as tokens, $S \in \mathbb{R}^{L \times D}$, where $L = (H/P) \times (W/P)$, with \textit{P} being the kernel size, denotes the sequence length, and \textit{D} is the vector dimension. In the encoding stage, a major fraction of the \textit{L} tokens are masked and removed, only the remaining tokens are used. The decoder receives the encoder's output, places them back in their original sequence positions, then fills in the masked positions with learnable mask tokens. The decoder generates a reconstructed image $\hat{I} \in \mathbb{R}^{C \times H \times W}$ with the same shape as the input image. $\hat{I}$ is compared to the original image using the mean-squared error (MSE) loss, calculated per-pixel exclusively on the masked tokens. Both the encoder and decoder are constructed using transformer blocks, though the encoder typically uses larger blocks. Fixed positional encoding is used to enable the model to capture the spatial information of tokens in the image.

\subsection{State Space Model}
The Mamba architecture is a new addition to State Space Models (SSM), representing a novel class of deep learning models inspired by continuous systems \cite{Gu2023,Dao2024}. They map a 1-D function or sequence \(x(t) \in \mathbb{R} \mapsto y(t) \in \mathbb{R}\) through a hidden state \(h(t) \in \mathbb{R}^{N}\).

\begin{equation} \label{eq2}
\begin{aligned}
  &h'(t) = Ah(t) + Bx(t), \\
  &y(t) = Ch(t) + Dx(t).
\end{aligned}
\end{equation}

where \(A \in \mathbb{R}^{N \times N}, B \in \mathbb{R}^{N \times 1}, C \in \mathbb{R}^{1 \times N}, D \in \mathbb{R}^{1 \times 1} \) are the learnable parameters. Mamba derives a discrete version of the continuous system using the zero-order hold (ZOH) rule, defined as follows:
\begin{equation} \label{eq3}
\begin{aligned}
  &\bar{A} = \exp(\Delta A), \\ 
  &\bar{B} = (\Delta A)^{-1} (\exp(\Delta A) - I) \cdot \Delta B.
\end{aligned}
\end{equation}	

where $\Delta$ is a timescale parameter. In the implementation, $\bar{B}$ is approximated as a first-order estimation, resulting in $\bar{B} = \Delta B$. Additionally, the three parameters $B$, $C$ and $\Delta$ are made selective by deriving directly from the input $x$ through learnable layers. Currently, there are two versions: Mamba 1 and Mamba 2. A key difference is that Mamba 2 features a multi-headed implementation with a shared timescale across the same-head features. Mamba 2 is significantly faster than Mamba 1 and is used in this research. 

The Mamba architecture was initially developed for language modeling. There are efforts to extend this architecture to the vision domain \cite{Liu2024,Zhu2024}. This involves unrolling and scanning an image in multiple directions, allowing the model to learn a more comprehensive representation at specific locations. For example, Vision Mamba and VMamba are two variations: Vision Mamba unrolls an image in row-major order and scans it both forward and backward, while VMamba extends the scanning to include column-major order.

\section{Method}
\label{sec:method}
\subsection{Pretraining method}
\subsubsection{Transformer-based architecture}
In recent years, advanced architectures like Prithvi and SpectralGPT have been developed for satellite imagery, focusing on input organization and masking strategies while retaining the MAE-ViT core structure. To enable a fundamental comparison between SatMamba and ViT-based methods, this study uses RGB images, common in high-resolution imagery. Future work will explore more complex domains, such as multispectral and multitemporal data. Therefore, the pretrain architecture is essentially similar to MAE \cite{He2022} for non-temporal RGB images. In the SatMAE study \cite{Cong2022}, researchers pretrained an MAE with a Vision Transformer Large (ViT-L) backbone (referred to here as ViTMAE-L) for high-resolution RGB images from the fMoW dataset. During the data preprocessing, the images from the fMoW dataset were cropped and resized, depending on the label object size in the image (see section ~\ref{sec:exppretraining} for detailed information about the fMoW dataset). Therefore, ViTMAE-L was pretrained on a resolution different from the actual resolution of this dataset. In this study, we pretrain a similar architecture on the fMoW dataset, but with images kept at their original resolution and using a Vision Transformer Base (ViT-B) backbone, referred to as ViTMAE-B.

\subsubsection{Mamba-based architecture}
\begin{figure}[ht]
    \centering
    \includegraphics[width=1.0\columnwidth]{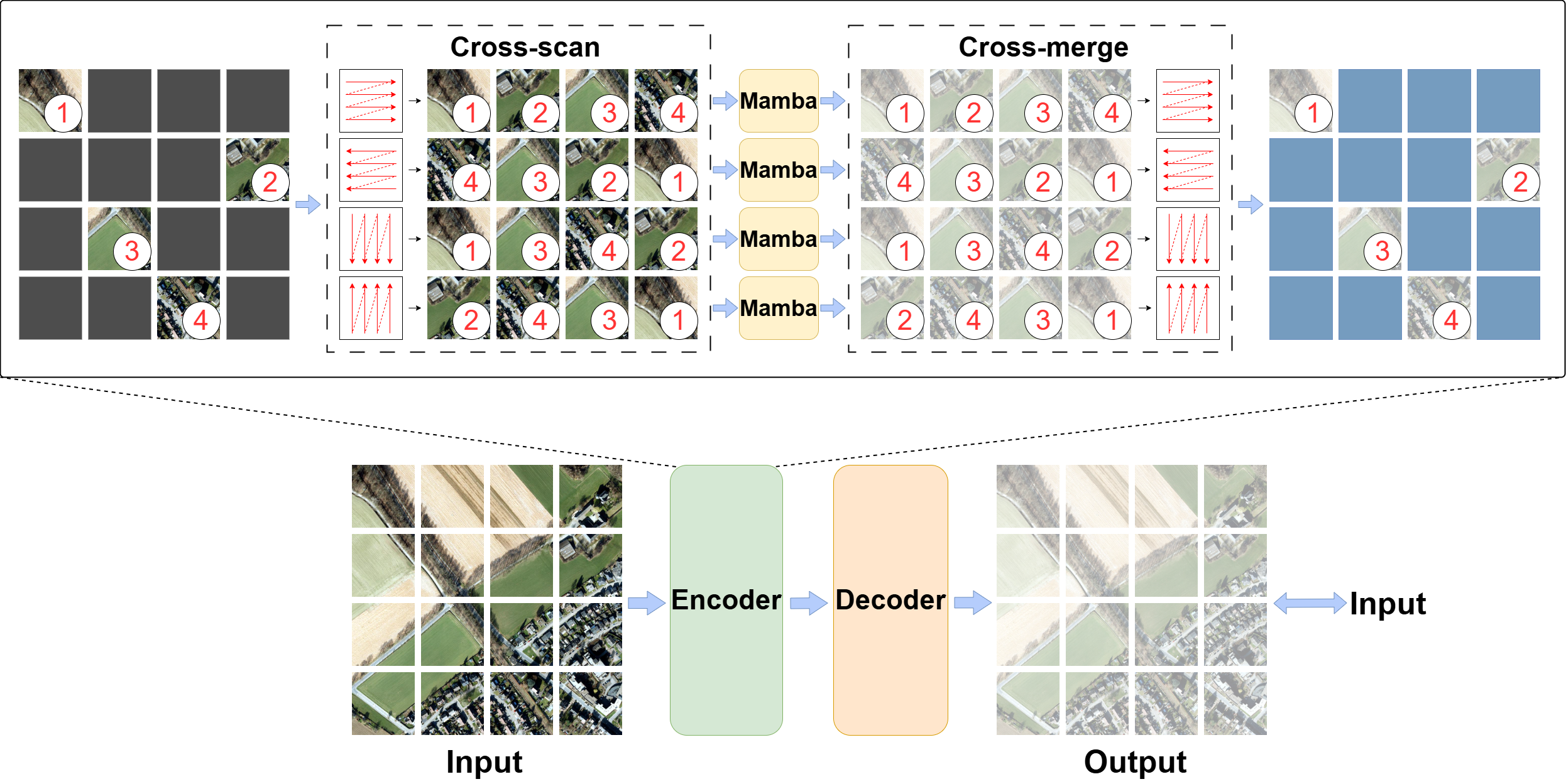}
    \caption{Details of the proposed SatMamba architecture. The details of the encoder are shown in the upper part of the figure. The decoder architecture is essentially similar to the encoder but smaller.}
    \label{fig:SatMamba}
\end{figure}

We introduced a pretrained architecture called SatMamba, as illustrated in Figure \ref{fig:SatMamba}. Given an input image of shape $\mathbb{R}^{C \times H \times W}$, we first patchify and apply linear projection to transform the input into a tensor of shape $\mathbb{R}^{(H/P) \times (H/P) \times D}$. This tensor is then fed to the masking procedure.
\paragraph{Masking}
Unlike MAE, where the input image is flattened in row-major order before masking, SatMamba first applies masking to the tensor. Our sampling strategy is random sampling without replacement. A certain masking ratio is applied to mask out a portion of the tensor. The masked tensor is then flattened before being fed to the encoder. Depending on the model's complexity, the tensor can be flattened in either row-major or column-major order, or both. According to the literature, multiple scanning directions can enhance the performance of Mamba-based vision models \cite{Liu2024, Zhu2024}. SatMamba supports up to four scanning directions: row-major forward and backward scanning, and column-major forward and backward scanning. Another key difference is that in SatMamba, the order of the flattened patches is maintained as they appear in the original tensor, unlike in MAE where patches are shuffled and positional encodings are used to infer the natural order. This design choice is made because Mamba is a sequence-based architecture that updates its internal state by sequentially processing observation inputs, so maintaining the natural order of patches is crucial for effective learning.
\paragraph{SatMamba Encoder}
Our encoder is composed of multiple layers, each being a multi-way Mamba block containing several Mamba blocks. The number of Mamba blocks within a multi-way Mamba block corresponds to the number of scanning directions, with each Mamba block responsible for a specific direction. In its full version, a multi-way Mamba block includes four Mamba blocks, each handling one of the following scanning directions: row-major forward, row-major backward, column-major forward, and column-major backward. The outputs of these blocks are then merged to produce the final output of the layer. 
\paragraph{SatMamba Decoder}
The output of the encoder is padded with mask tokens and then reordered to its original sequence. The decoder consists of another series of multi-way Mamba blocks and is designed to be smaller than the encoder, with its depth and embedding size similar to that of MAE's decoder.

\paragraph{Reconstruction target}
Naturally, the reconstruction target is the original pixel values. Another variant is the normalized pixel values of each masked patch, which has been shown to improve the network's representation quality \cite{He2022}. Due to limited computational resources, we use this variant as the reconstruction target in all of our experiments.

\paragraph{Positional encodings}
Positional encodings provide information about the structure and position of data in ViT. However, the importance of positional encodings in SatMamba still requires investigation. Therefore, we experimented with both versions. Fixed positional encodings, similar to those used in MAE, are applied.

\subsection{Fine-tuning method}
We used the following backbones for comparison: ResNet50, EfficientNet-B7, ViTMAE-L, ViTMAE-B, and SatMamba-B. ResNet50 is commonly used as a baseline for evaluating foundation models in remote sensing. In addition, EfficientNet-B7, the largest model in the scalable CNN-based EfficientNet architecture family, also serves as a strong baseline for comparison with foundation models. This study uses two datasets: OpenEarthMap for land cover classification and xBD for assessing building damage after natural disasters. The first task, known as semantic segmentation, involves assigning semantic labels to each pixel in an image, corresponding to land cover classes as defined by the dataset. The dataset for this task can be denoted as $\{(x_i, y_i)\}_{\scriptstyle i=1}^{N}$, where $y_i \in \{0, 1, ..., C\}^{H \times W}$ represents the labeled image. For this task, we base our architectures on UNet \cite{Ronneberger2015}, customizing them according to the specific backbone used. The UNet implementation follows that of \cite{Iakubovskii2019}, with a depth of 5. For the MAE architectures using transformers or mamba as proposed, the depth is set to 4 to accommodate a patch size of 16. The UNet configuration for ViTMAE and SatMamba is adapted from \cite{IsaacCorley2024}. 

The building damage assessment task involves change detection using pairs of pre and post-disaster images. The dataset for this task can be denoted as $\{(x_i^{t_{1}}, x_i^{t_{2}}, y_i^{loc}, y_i^{clf})\}_{\scriptstyle i=1}^{N}$, where $y_i^{loc} \in \{0, 1\}^{H \times W}$ represents the building segmentation label at time $t_{1}$ (pre-disaster), and $y_i^{clf} \in \{0, 1, ..., C\}^{H \times W}$ represents the damage classes at time $t_{2}$ (post-disaster). We design an architecture based on UNet that enables joint learning of multiple tasks. We follow the protocol of the xView2 Challenge \cite{Gupta_2019_CVPR_Workshops} to split this task into two subtasks: building localization and damage classification. The model's input consists of two images: before and after the disaster. Both images are processed through the same shared encoder. The features from the pre-disaster image are then used as input for a decoder specifically for building localization. Simultaneously, in the damage assessment branch, features from both the pre- and post-disaster images are concatenated and fed into a separate decoder for damage assessment. Essentially, this is an extended version of the semantic segmentation architecture mentioned above, with an additional branch dedicated to damage assessment. All of these outputs are combined through a single loss function.
In the evaluation phase, both models are post-processed using a method that produces the final output by computing a weighted average of four different predictions. These predictions are generated by applying the model to transformations of the original input in the spatial dimensions.

\section{Experiment}
\label{sec:experiment}
\subsection{Pretraining}
\label{sec:exppretraining}
The Functional Map of the World (fMoW) dataset \cite{Christie2018} is used for pretraining. Out of a total of 416,614 RGB images, 363,572 were used for training and 53,042 for validation. The training strategy is similar to those used in the SatMAE and MAE studies. The input images are 224x224 pixels with a patch size of 16x16 pixels. For each image, 75\% of all patches are masked. We used an independent masking strategy where the masked regions vary across images. The reconstruction target is the normalized pixel values of each masked patch, aiming for better representation quality. The training was conducted on an NVIDIA RTX 6000 Ada, with the batch size optimized to fully utilize the GPU memory. Training parameters, image size (224x224), the optimizer, and the learning rate scheduler were set similarly to \cite{Cong2022}. 

\begin{figure}[!t]
  \centering
  \begin{tabular}{@{}c@{\hspace{0.1cm}}c@{}} 
    \includegraphics[width=0.5\columnwidth]{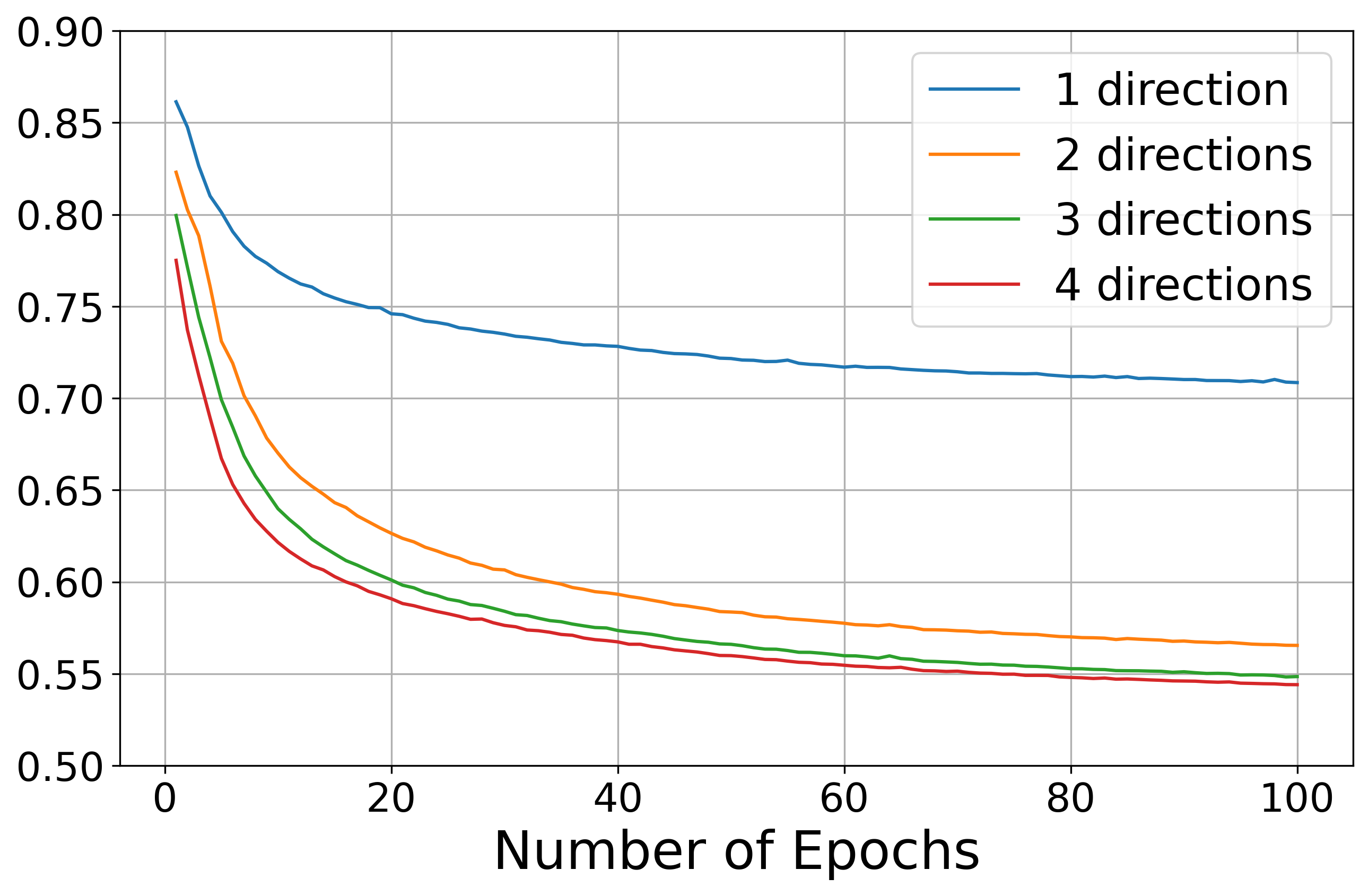} &  
    \includegraphics[width=0.5\columnwidth]{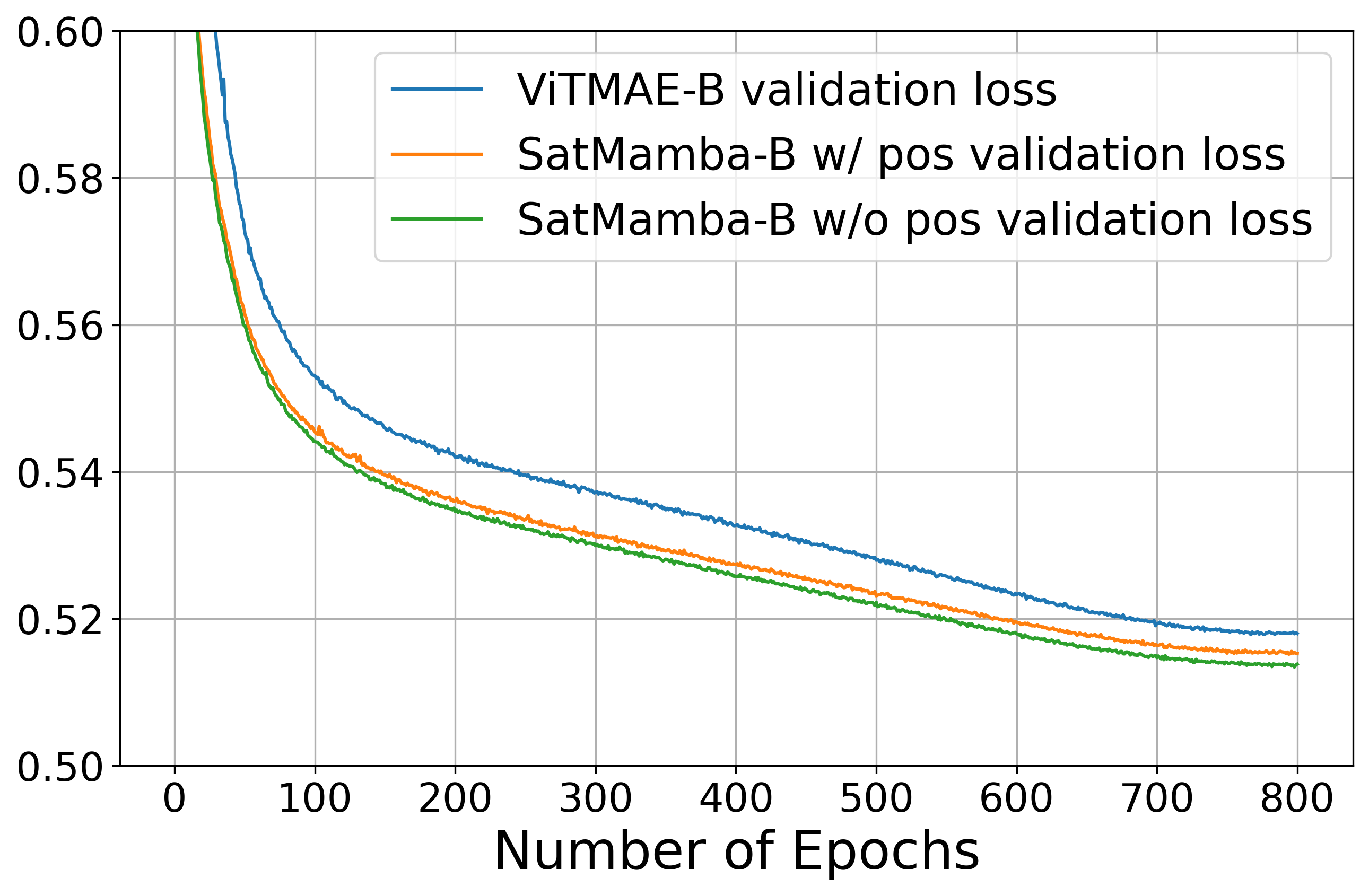} \\ 
    (a) & (b) 
  \end{tabular}
  \caption{Pretraining results of SatMamba-B and ViTMAE-B on the fMoW dataset: (a) Ablation experiments with different scanning directions over 100 epochs; (b) Full pretraining results of SatMamba-B using four scanning directions over 800 epochs.}
  \label{fig:trainingResult}
\end{figure}

\begin{figure}[!t]
  \centering
  \begin{tabular}{@{}c@{\hspace{0.1cm}}c@{}} 
    \includegraphics[width=0.5\columnwidth]{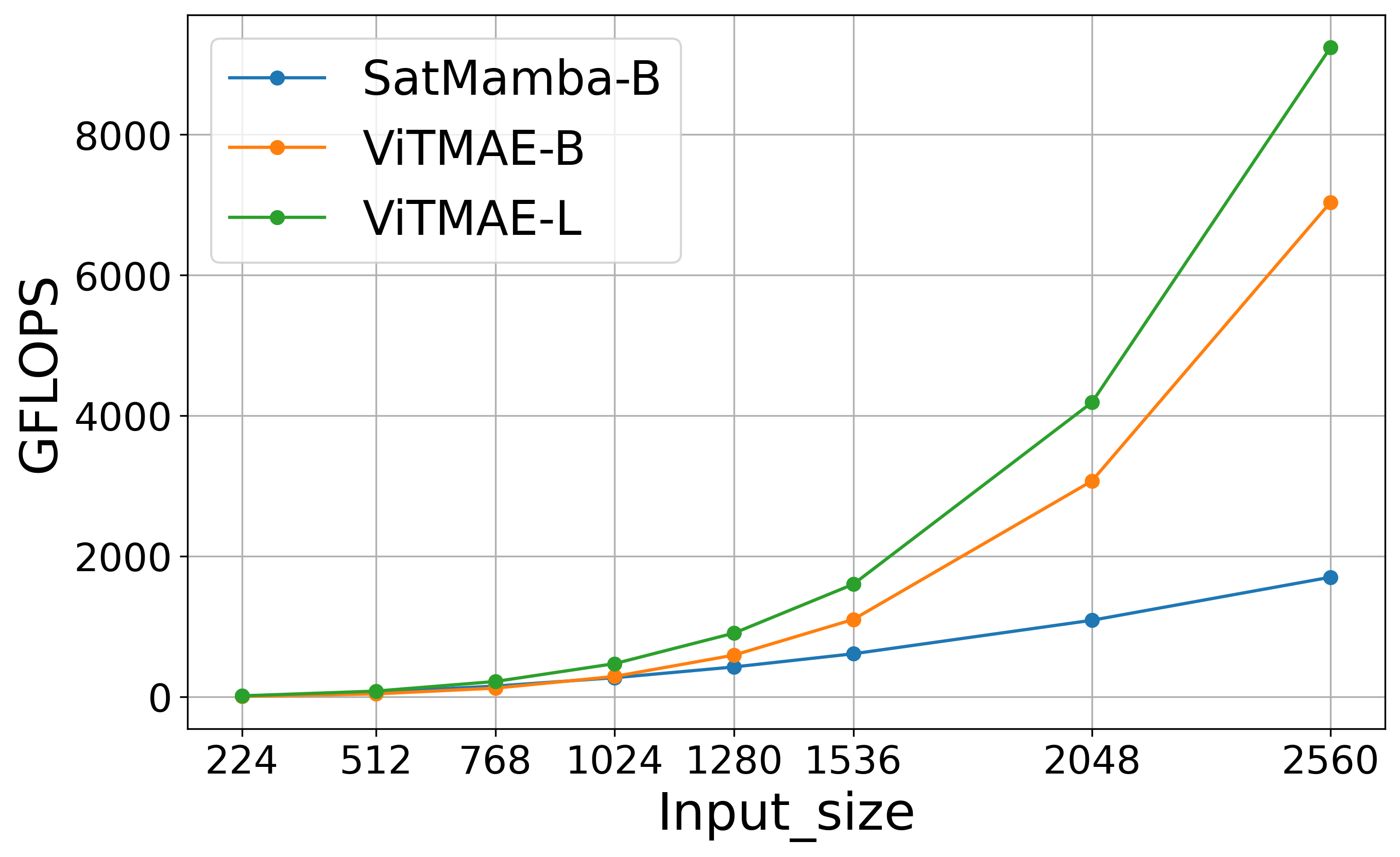} &  
    \includegraphics[width=0.5\columnwidth]{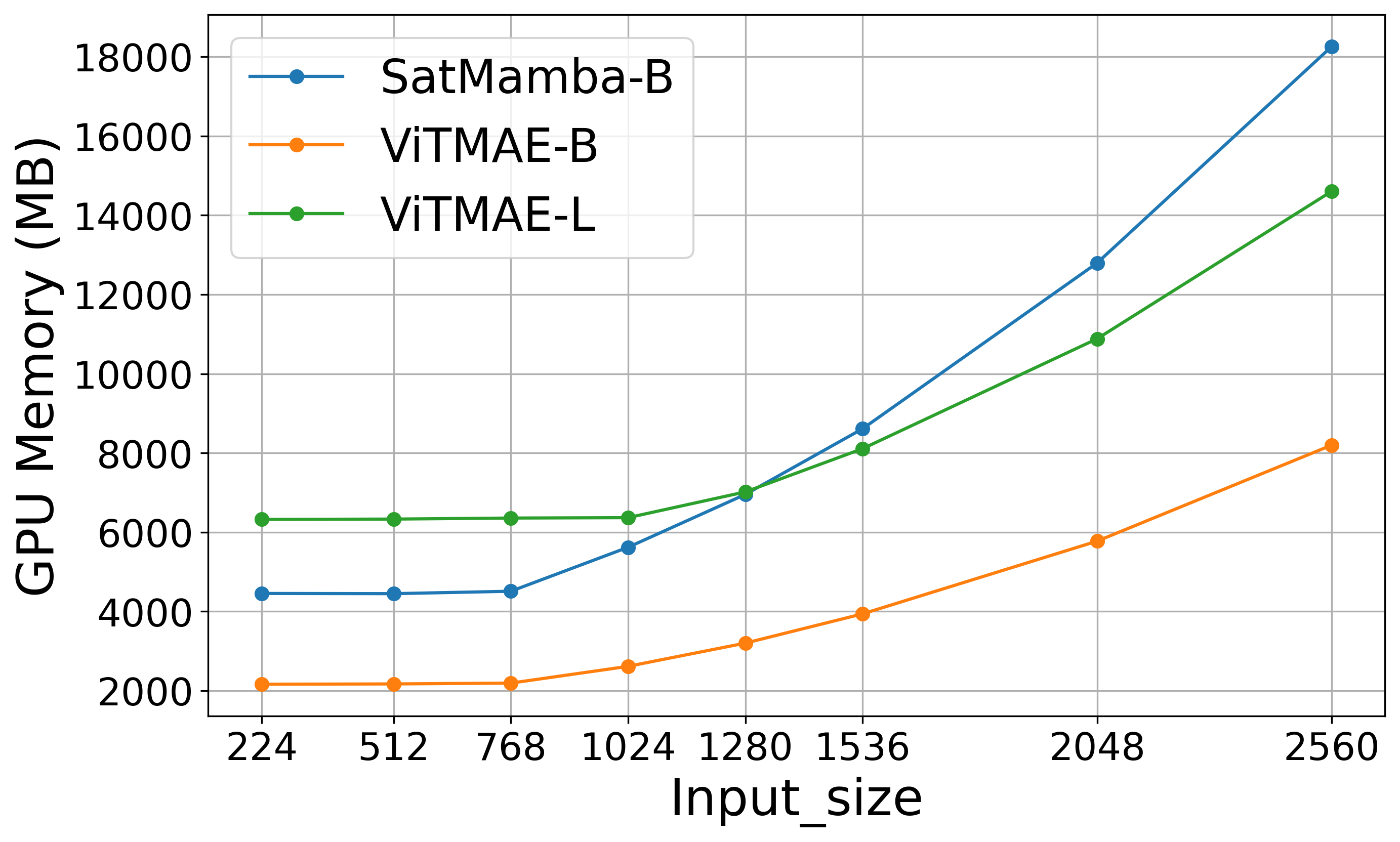} \\ 
  \end{tabular}
  \caption{ \small Resource requirements of different models across varying input sizes: (a) Computational requirements; (b) Memory requirements.}
  \label{fig:resource}
\end{figure}

Figure \ref{fig:trainingResult} shows the pretraining results. The ablation experiments demonstrated that SatMamba significantly benefits from multiple scanning directions, with the lowest loss observed in full scanning. During full training, both versions of SatMamba-B, with (SatMamba-B w/ pos) and without positional encodings (SatMamba-B w/o pos), exhibited lower training and validation losses compared to ViTMAE-B. Additionally, SatMamba-B w/o pos outperformed SatMamba-B w/ pos, with the validation loss further diverging toward the end of the training. We configured SatMamba-B to align with the settings of ViTMAE-B (Table \ref{tab:SettingParams}). Additionally, the Mamba-specific parameters, including state dimension and head dimension, are set to 64 and 96, respectively. In the full scan, each layer of SatMamba-B consists of 4 Mamba blocks. Therefore, although each Mamba block contains about half the parameters of a ViT block, each SatMamba-B layer has roughly twice the parameters of a ViTMAE-B layer. This increased parameter count was necessary to achieve pretraining results comparable to those of ViTMAE-B. It is important to note that these results are meant for baseline comparison, highlighting that the two models exhibit comparable pretraining performance. However, it is not necessary that the pretraining loss transfers linearly to the fine-tuning loss.
\begin{table}[t]
	\caption{Configuration of experimented foundation models. The SatMamba configuration is shown in the four-scanning model.}
  \label{tab:SettingParams}
	\centering
  \resizebox{\columnwidth}{!}{%
	\begin{tabular}{l|l|l|l}
		Config & SatMamba-B & ViTMAE-B & ViTMAE-L \\
		\hline
		Encoder embedding dimension & 768 & 768 & 1024 \\
		Encoder depth & 12 & 12 & 24 \\
    Encoder number of heads &  & 12 & 16 \\
		Decoder embedding dimension & 512 & 512 & 512 \\
    Decoder depth & 8 & 8 & 8 \\
    Decoder number of heads & & 16 & 16 \\
    Inner state dimension & 64 & & \\
    Head dimension & 96 & & \\
    \hline
    Param (M) & 229.86 & 111.66 & 329.24 \\
	\end{tabular}
  }
\end{table}
The computational cost of SatMamba increases linearly with input size (Figure \ref{fig:resource}). This property is particularly advantageous when handling long input sequences, a common scenario in remote sensing, especially with multispectral, multitemporal, and hyperspectral data. Currently, RS foundation models developed with ViT face challenges due to its quadratic scaling with sequence length, leading to the adoption of various strategies to mitigate this issue. These include spectral group patching/pooling, where specific channel groups, such as blue-green-red bands, are embedded or pooled together \cite{Cong2022,Irvin2023}. While these techniques offer certain advantages, they may require a trade-off between performance and sequence length. SatMamba, on the other hand, allows more flexible use of spectral channels, sensors, and temporal lengths. However, it incurs a higher initial computational cost at smaller input sizes or shorter sequence lengths, such as 224x224. Specifically, at this resolution, SatMamba's computational cost is twice that of ViTMAE-B and slightly less than ViTMAE-L. Additionally, the current implementation of SatMamba demands more GPU memory than ViTMAE-B for smaller input sizes, and exceeds the memory requirements of both ViTMAE versions as input size increases. Future research is expected to improve the model's memory usage, particularly by developing better scanning approaches and kernels.

\subsection{Downstream Tasks}

\subsubsection{Semantic Segmentation} In this experiment, we use the OpenEarthMap dataset \cite{Xia2023}, which consists of 5,000 high-resolution aerial and satellite images from 44 countries. Each image is manually annotated with 8 land cover labels: bare land, rangeland, developed space, road, tree, water, agricultural land, and building. The downloadable dataset contains 3,500 images, consisting of 3,000 training images and 500 validation images. In all experiments with this dataset, we split the original training images into two parts: 2,500 images for training and 500 images for validation during the training process. The original validation dataset is then used as the test set. All experiments used the AdamW optimizer with a warmup and half-cycle cosine decay schedule, along with tuned learning rates. Training is conducted for 100 epochs with an early stopping strategy. Other training parameters are set as in \cite{He2022}. 

\begin{table}[!t]
	\caption{Results of semantic segmentation on the OpenEarthMap dataset. $\dagger$ denotes supervised training from scratch}
  \label{tab:oedResult}
	\centering
	\resizebox{\columnwidth}{!}{
	\begin{tabular}{llllllllll}
		\toprule
		\midrule
		\multicolumn{1}{c}{} & \multicolumn{8}{c}{IoU (\%)} & \multicolumn{1}{c}{} \\
		\cmidrule(r){2-9}
		Backbone & Bareland & Rangeland & Developed & Road & Tree & Water & Agriculture & Building & mIoU (\%) \\
		\midrule
    ResNet50 & 37.40 & 55.99 & 55.68 & 65.12 & 70.93 & 76.70 & 75.17 & 78.81 & 64.48 \\
		EfficientNet-B7 & 39.20 & \textbf{58.91} & \textbf{58.02} & \textbf{66.25} & 70.03 & 77.30 & \textbf{79.19} & 79.83 & 66.34 \\
		\midrule
    	$\text{ViTMAE-B}^\dagger$ & 29.03 & 48.56 & 45.35 & 47.90 & 64.93 & 69.56 & 64.07 & 67.18 & 54.57 \\
		ViTMAE-B & 38.45 & 57.07 & 56.87 & 65.80 & 71.63 & \textbf{79.33} & 75.45 & \textbf{80.40} & 65.62 \\
    	$\text{ViTMAE-L}^\dagger$ & 30.73 & 49.68 & 46.36 & 45.78 & 64.51 & 68.42 & 65.24 & 67.20 & 54.74 \\
		ViTMAE-L & 39.32 & 55.76 & 56.25 & 62.20 & 71.97 & 78.35 & 75.48 & 79.14 & 64.81 \\
		\midrule
		SatMamba-B w/ pos & 41.19 & 57.12 & 54.72 & 63.56 & \textbf{72.22} & 76.03 & 74.02 & 78.79 & 64.71 \\
		$\text{SatMamba-B w/o pos}^\dagger$ & 39.26 & 52.99 & 52.55 & 60.82 & 66.43 & 63.93 & 71.16 & 75.21 & 60.29 \\
    SatMamba-B w/o pos & \textbf{43.44} & 58.79 & 55.39 & 65.98 & 72.17 & 78.83 & 77.54 & 79.57 & \textbf{66.46} \\
		\midrule
		\bottomrule
	\end{tabular}
	}
\end{table}

We calculated Intersection over Union (IoU) for each class, and the mean IoU (mIoU) is used as an overall performance evaluation metric (Table \ref{tab:oedResult}). All experiments using pretrained backbones significantly outperformed those trained from scratch, particularly with ViT-based models and, to a lesser extent, with SatMamba. The SatMamba-B w/o pos model achieved the highest mIoU of 66.46\%, demonstrating that the Mamba block can be effectively integrated into the MAE framework. Additionally, SatMamba-B w/o pos outperformed SatMamba-B w/ pos by 1.75\%, suggesting that positional encodings can influence the model's performance. The CNN-based models also showed competitive performance, with EfficientNet-B7 achieving an mIoU of 66.34\%. This finding aligns with recent studies suggesting that CNN models can deliver results comparable to those of transformer-based foundation models on remote sensing tasks when using similar preprocessing steps as the pretraining phase \cite{Corley2023, Fibaek2024}. Interestingly, within the ViT-based models, ViTMAE-B outperformed ViTMAE-L by 0.81\%, despite being a smaller version. Several factors may explain this difference, including pretraining resolution and model size. ViTMAE-B was pretrained at the original resolution, while ViTMAE-L used a modified resolution, which may have impacted its fine-tuning performance. Furthermore, ViTMAE-L, being over three times larger than ViTMAE-B, likely required more pretraining data, which could have contributed to its relatively lower performance.

\subsubsection{Building Damage Assessment} In this experiment, we use the xBD dataset \cite{Gupta_2019_CVPR_Workshops}, a global dataset for building damage assessment. It contains a total of 11,034 pairs of pre- and post-disaster images for 19 disaster events of 6 types worldwide. The dataset is divided into training, validation, and test sets, with 9168, 933, and 933 pairs, respectively. For each pair of images, building footprints are labeled in both. Additionally, in the post-disaster images, each building is further annotated with a damage level, which can be one of four levels: no damage, minor damage, major damage, or destroyed. Training parameters are implemented as in the semantic segmentation experiment.

\begin{table}[!t]
  \caption{Results of building damage assessment on the xBD dataset. $\dagger$ denotes supervised training from scratch.}
  \label{tab:xBDResult}
  \centering
  \resizebox{\columnwidth}{!}{
      \begin{tabular}{lccccccccc}
          \toprule
          \midrule
          \multicolumn{4}{c}{} & \multicolumn{4}{c}{Damage $F_{1}$ per class} & \multicolumn{1}{c}{} \\
          \cmidrule(r){6-9}
          Backbone & IoU & $F_{1}^{loc}$ & $F_{1}^{clf}$ & $F_{1}^{overall}$ & No damage & Minor damage & Major damage & Destroyed \\
          \midrule
          ResNet50 & 75.35 & 85.94 & 75.76 & 78.81 & 94.19 & 58.95 & 74.33 & 84.97 \\
          EfficientNet-B7 & 75.70 & 86.17 & 77.46 & 80.08 & 94.58 & 61.51 & \textbf{76.32} & 85.44 \\
		  \midrule
		  $\text{ViTMAE-B}^\dagger$ & 64.40 & 78.34 & 72.41 & 74.19 & 92.49 & 56.33 & 69.67 & 81.18 \\
          ViTMAE-B & 76.03 & 86.38 & 76.47 & 79.45 & 94.34 & 59.54 & 75.61 & 85.58  \\
		  $\text{ViTMAE-L}^\dagger$ & 61.07 & 75.83 & 70.79 & 72.31 & 92.19 & 52.71 & 69.43 & 81.43 \\
          ViTMAE-L & 73.52 & 84.74 & 73.47 & 76.85 & 93.14 & 55.66 & 72.88 & 83.18  \\
		  \midrule
          SatMamba-B w/ pos & 75.72 & 86.18 & 76.15 & 79.16 & 94.22 & 59.48 & 73.95 & 86.35 \\
		  $\text{SatMamba-B w/o pos}^\dagger$ & 71.06 & 83.08 & 75.51 & 77.78 & 93.66 & 59.41 & 73.64 & 84.12 \\
          SatMamba-B w/o pos & \textbf{77.04} & \textbf{87.03} & \textbf{77.77} & \textbf{80.55} & \textbf{94.80} & \textbf{61.84} & 76.03 & \textbf{86.51} \\
          \midrule
          \bottomrule
      \end{tabular}
  }
\end{table}

For evaluation metrics, we follow the protocol of \cite{Chen2024}. Table \ref{tab:xBDResult} presents the results of the models. The fine-tuned models using pretrained backbones significantly outperformed their supervised training from scratch counterparts. In this experiment, SatMamba-B w/o pos achieved the highest overall F1 score at 80.55\%, with top F1 scores in three damage categories: no damage (94.80\%), minor damage (61.84\%), and destroyed (86.51\%). SatMamba-B w/o pos outperformed SatMamba-B w pos by a margin of 1.39\% in $F_{1}^{overall}$. EfficientNet-B7 achieved the second-highest overall F1 score of $F_{1}^{overall}$ of 80.08\% with the top F1 score of 76.32\% for the major damage class. The experiment further confirms the superior performance of ViTMAE-B compared to ViTMAE-L, despite ViTMAE-L having over three times more parameters. ViTMAE-L's overall F1 score was significantly lower by 2.6\% compared to the ViTMAE-B model. 

\FloatBarrier

\section{Conclusion}
\label{sec:conclusion}

This paper introduced the SatMamba architecture to enable efficient self-supervised learning. The pretrained SatMamba-B demonstrated competitive pretraining performance compared to ViT-based methods, with lower computational costs as input size increases. However, it incurs high initial costs. Future work is expected to reduce the computational costs and memory requirements of the model. In experiments on two image-to-image tasks, SatMamba delivered competitive results against strong benchmarks such as ResNet50, EfficientNet-B7, ViTMAE-B, ViTMAE-L. Several promising avenues for future development include scaling to stronger models by utilizing larger pretraining datasets, as well as experimenting with SatMamba in other image domains. The scaling capabilities of Mamba-based architectures as model size grows, compared to ViT-based architectures, is an active research area in both remote sensing and other domains. Additionally, since SatMamba can be adapted to other image domains, such as medium-resolution, multispectral RS and natural imageries, it would be worthwhile to test the architecture in these areas as well. 

\section*{Acknowledgments}
This work was supported by the International Digital Earth Applied Science Research Center, an International Joint Usage/Research Center located at Chubu University, Japan.

\bibliographystyle{IEEEtran}
\bibliography{references}

\end{document}